\documentclass[10pt,twocolumn,letterpaper]{article}

\usepackage{cvpr}              
\usepackage{graphicx}
\usepackage{amsmath}
\usepackage{amssymb}
\usepackage{booktabs}
\usepackage{enumitem}
\usepackage{multirow}
\usepackage[nodisplayskipstretch]{setspace}
\usepackage{array, multirow}
\usepackage{amsmath,amssymb,stmaryrd}   
\usepackage{multicol}
\usepackage[pagebackref,breaklinks,colorlinks]{hyperref}
\usepackage[capitalize]{cleveref}
\crefname{section}{Sec.}{Secs.}
\Crefname{section}{Section}{Sections}
\Crefname{table}{Table}{Tables}
\crefname{table}{Tab.}{Tabs.}
\def\cvprPaperID{21} 
\def\confName{CVPR}
\def\confYear{2022}

\newcommand{\comment}[1]{}
\begin{document}

\title{Latents2Segments: Disentangling the Latent Space of Generative Models for Semantic Segmentation of Face Images}

\author{Snehal Singh Tomar \quad A.N. Rajagopalan\\
Indian Institute of Technology Madras\\
{\tt\small snehal@smail.iitm.ac.in, raju@ee.iitm.ac.in}
}
\maketitle

\begin{abstract}
 With the advent of an increasing number of Augmented and Virtual Reality applications that aim to perform meaningful and controlled style edits on images of human faces, the impetus for the task of parsing face images to produce accurate and fine-grained semantic segmentation maps is more than ever before. Few State of the Art (SOTA) methods which solve this problem, do so by incorporating priors with respect to facial structure or other face attributes such as expression and pose in their deep classifier architecture. Our endeavour in this work is to do away with the priors and complex pre-processing operations required by SOTA multi-class face segmentation models by reframing this operation as a downstream task post infusion of disentanglement with respect to facial semantic regions of interest (ROIs) in the latent space of a Generative Autoencoder model. We present results for our model's performance on the CelebAMask-HQ and HELEN datasets. The encoded latent space of our model achieves significantly higher disentanglement with respect to semantic ROIs than that of other SOTA works. Moreover, it achieves a 13\% faster inference rate and comparable accuracy with respect to the publicly available SOTA for the downstream task of semantic segmentation of face images.               
\end{abstract}
\vspace{-5mm}
\section{Introduction}
\label{sec:intro}
\setlength{\tabcolsep}{3pt}
Multi-class semantic segmentation of facial regions of interest is central to various AR and VR applications. Yet, there exists a paucity of publicly available pre-trained models which can perform this task with reasonable accuracy.
\begin{figure}[h]
\centering
\includegraphics[scale=0.46]{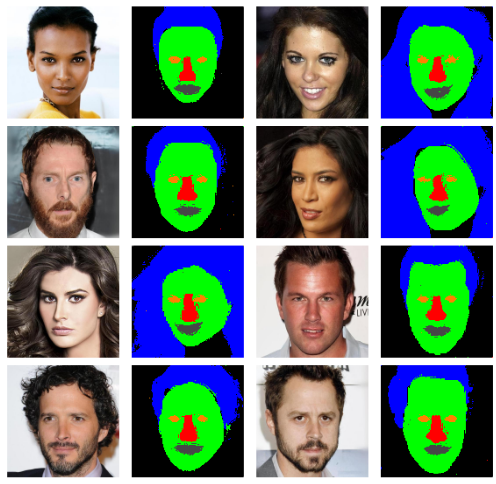}
\caption{Representative semantic segmentation results on test images sampled from the CelebAMask-HQ dataset. Columns 1, 3 contain input face images and Columns 2, 4 contain their corresponding output segmentation maps, respectively. The color coding used for semantic regions in the segmentation maps is given by; blue: \textit{hair}, green: \textit{skin}, red: \textit{nose}, orange: \textit{eyes}, and grey: \textit{lips + mouth}.}
\end{figure}
The promise of deep learning for general semantic segmentation has been explored by several works (\cite{DBLP:journals/corr/RonnebergerFB15}, \cite{He_2017_ICCV}, \cite{8578572}, \cite{DBLP:journals/corr/abs-1804-02767}, \cite{Liu_2019_CVPR},  \cite{Aygun_2021_CVPR}) across a variety of scene settings. Face segmentation is a particularly challenging problem because of the irregular shapes, sizes, and textures of facial regions of interest.
\begin{figure*}
  \centering
  \includegraphics[scale=0.48]{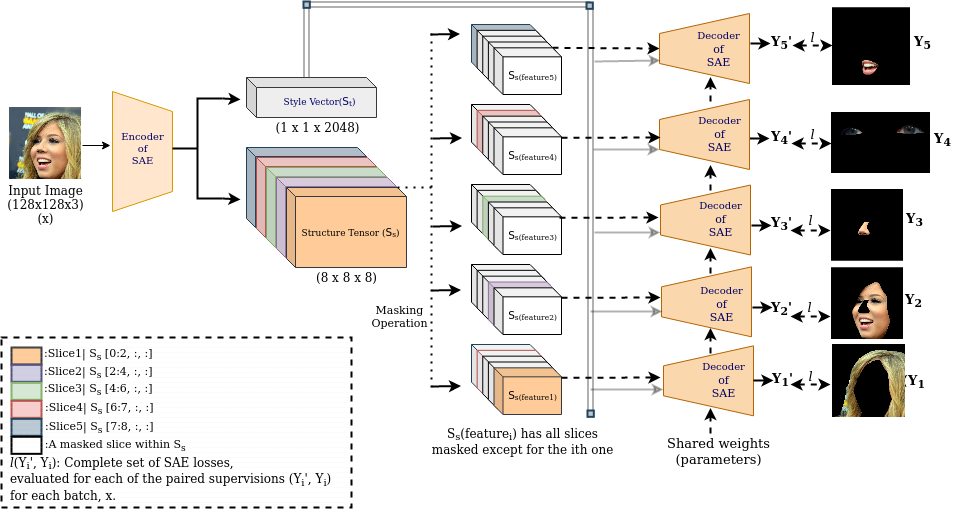}
    \caption{A schematic representation of our model's architecture and training pipeline. The schema used for slicing $S_{s}$ and for formation of masked $S_{s_{feature_{i}}}$ have been annotated in the legend. $l$ refers to the operation defined by Eq. 1.}  
\end{figure*}
A category of literature (\cite{DBLP:journals/corr/abs-1907-06740}, 
\cite{DBLP:journals/corr/abs-1712-07168}) solves the problem of segmenting out one region of interest (hair in most cases) by incorporating 
priors unique to that region in their architecture and losses. The SOTA when it comes to segmenting multiple regions of interest in face images are \cite{Lin_2019_CVPR} and \cite{te2020edge}. While, \cite{Lin_2019_CVPR} relies on heavily pre-processed images, \cite{te2020edge} incorporates relationships with facial expressions by learning graph representations. The computational overhead that pre-processing operations (warping) incur and the narrow scope of generalization of representation learning are key limitations of these works. These bottlenecks warrant the need for a pre-processing and prior independent approach that can generate fine-grained multi-class segmentation maps with a single forward pass over the input images. To this end, we propose the use of Generative Autoencoders (GAs) capable of producing regions of interest in a selective fashion, given an input image. GA models have shown promising results for tasks such as high fidelity reconstruction of images, style transfer, and style manipulation. However, the latent space of these models is heavily entangled and very high dimensional in nature. This limits their ability to enable spatially localised manipulation of images as a consequence of specific perturbations to their encoded latent space. The Swapping Autoencoder (SAE) \cite{park2020swapping} is an especially insightful work as it learns a latent representation with a neat distinction between structure and style information of the input image. Inspired by \cite{park2020swapping}, we build upon its capabilities further in this work.
Previously, generative models have been used extensively (\cite{Isola_2017_CVPR}, \cite{Zhu_2017_ICCV}, \cite{Park_2019_CVPR},\cite{Tan_2021_CVPR}, \cite{DBLP:journals/corr/abs-2012-04644}) for semantic image synthesis, which is the task of generating photo-realistic images that match the structure of a semantic segmentation map given as input. However, the problem at hand has not seen  significant attempts in the past. The contributions of this paper can be summarized as below:
\\
\vspace{-6.5mm}
\\
\begin{itemize}[noitemsep, nolistsep]
    \item We infuse a strong disentanglement with respect to the structure of semantic ROIs in re-generated images, in the latent space of 
    SAE \cite{park2020swapping}. Ours is the first work which does so for any Generative Autoencoder model. We provide quantitative metrics for the extent of disentanglement achieved.
    \item We harness the disentangled nature of our model's latent space to perform the challenging downstream task of semantic face segmentation. Results obtained for this task are close to the current SOTA.
    \item Generating the segmentation map for a given ROI from an image belonging to a particular distribution using our model trained on a similar distribution amounts to a simple forward pass with appropriate masking (retention of a single non-zero slice corresponding to the ROI) applied to the latent space. This eliminates the need of any priors for semantic segmentation and underscores the applicability of our model to any generic semantic segmentation problem.  
\end{itemize}
\section{Methodology}
The SAE\cite{park2020swapping} is a generative Autoencoder model which embeds the \textit{structure} and \textit{style} information present in input images ($H \times H \times 3$) into a \textit{structure tensor} ($S_{s}$, having dimensions $H/16 \times H/16  \times 8$) and \textit{texture vector} ($S_{t}$, having dimensions $1 \times 1 \times 2048$). The latent space $S = \{S_{s}, S_{t}\}$ serves as the point of initiation for our work. Our objective in this work, is to disentangle the tensor slices within $S_{s}$ such that they correspond to the structure information of individual regions of interest, namely: hair, skin, nose, eyes, and (lips + mouth) in the reconstructed image. This, in effect, entails that \textit{masking (setting to zero) all slices of $S_{s}$ except one should should produce an image containing only its corresponding semantic region of interest, when decoded together with $S_{t}$}. This observation is key to our work and the results and experiments that follow, are based on it.
\subsection{Network Architecture and Losses}
 Figure 2 depicts our model which is a parallelized version of the SAE\cite{park2020swapping} wherein, disentanglement with respect to semantic regions of interest is infused in the latent space. We achieve this by slicing the structure tensor and enforcing faithful reconstruction of the ROI by the decoder, when given the texture vector and a masked structure tensor (having only one nonzero slice that should correspond to the chosen ROI) as input. 
The correspondences that we have sought to develop are; Slice 1 (Channels 1 and 2 of $S_{s}$): \textbf{\textit{Hair}} ($R_{1}$), Slice 2 (Channels 3 and 4 of $S_{s}$): \textbf{\textit{Skin}} ($R_{2}$), Slice 3 (Channels 5 and 6 of $S_{s}$): \textbf{\textit{Nose}} ($R_{3}$), Slice 4 (Channel 7 of $S_{s}$): \textbf{\textit{Eyes}} ($R_{4}$), and Slice 5 (Channel 8 of $S_{s}$): \textbf{\textit{Lips + Mouth}} ($R_{5}$). We did not optimize the chosen slicing scheme on the basis of amount of disentanglement obtained. Thus, our results are independent of the chosen slicing scheme. We treat each Siamese decoder together with the encoder as a separate SAE (including all its components viz. the encoder, generator (decoder in our case), discriminator, and patch co-occurrence discriminator) while computing losses. The loss for each correspondence pair is defined as:
\begin{equation}
\begin{split}
      l(Y_{i}', Y_{i}) = L_{\text{rec}}(Y_{i}', Y_{i}) & + 0.5L_{\text{GAN, rec}}(Y_{i}', Y_{i})\\ +  0.5L_{\text{GAN, swap}}(Y_{i}', Y_{i}) &  + 0.5L_{\text{CooccurGAN}}(Y_{i}', Y_{i})
\end{split}
\end{equation}
Here, $Y_{i}'$ referes to the reconstruction obtained from the decoder for the latent space $\{S_{s_{feature_{i}}}, S_{t}\}$, $Y_{i}$ refers to the ground-truth image containing $R_{i}$ only, $L_{\text{rec}}$ refers to the reconstruction(L1) loss, $L_{\text{GAN, rec}}$ refers to the non-saturating GAN loss, $L_{\text{GAN, swap}}$ refers to the non-saturating GAN loss for images generated post swapping, and $L_{\text{CooccurGAN}}$ refers to the patch co-occurrence discriminator loss as defined in \cite{park2020swapping}. For every batch of training data, parameters of the encoder and decoder (all Siamese decoders share the same parameters) are optimized using the following overall loss:
\begin{equation}
    L_{\text{overall}} = \sum_{i = 1}^{5} 0.2 \cdot l(Y_{i}', Y_{i})
\end{equation}
Training was initiated using the pre-trained weights provided by\cite{park2020swapping}, post training on the FFHQ dataset\cite{DBLP:journals/corr/abs-1812-04948}. Inferring segmentation maps from our model amounts to a simple forward pass over the network with appropriate masking applied to $S_{s}$ in the latent space, depending upon the ROI for which the segmentation has to be performed. 
\subsection{Data Preparation}
We have used images (resized to $128\times128\times3$ dimensions) from the CelebAMask-HQ\cite{CelebAMask-HQ} and HELEN\cite{10.1007/978-3-642-33712-3_49} (contains images in the wild) datasets for training and evaluation. A batch of training data comprised of $ \{X, Y_{1}, ..., Y_{5}\}$ where $X$ denotes a batch of input images and $Y_{i}$ denotes a batch of region specific images, $R_{i}$. We sampled only those images from the CelebAMask-HQ dataset which had annotations for all the intended ROIs and split them into 27016 training and 862 test images. The HELEN dataset was used as made available.    
\section{Experiments}
Since the objective of this work is to disentangle the latent space of a GA model so as to develop strong correspondences between latent slices and semantic regions of interest in order to harness the same for the downstream application of semantic segmentation of face images, we analyse our model's performance primarily with respect to two criteria, namely the \textit{Degree of Disentanglement} achieved and the \textit{accuracy} of predictions for the chosen downstream task. 
\subsection{Degree of Disentanglement}
Degree of Disentanglement implies the extent to which changes made within a chosen slice of the latent representation ($S_{s}$) affect the pixel intensities in the desired semantic region of interest ($R_{i}$),  without causing any deviation to those in other ROIs. We propose the use of an activeness
measure inspired by \cite{peebles2020hessian} to quantify the degree to which a latent space conforms to this desired attribute. We define the activeness of a latent tensor masked in accordance with the $i^{th}$  slice of $S_{s}$ and denoted as  $S_{s_{feature_{i}}}$, with respect to a particular semantic ROI ($R_{j}$) as:
\begin{equation}
    A_{ij} = \mathbb{E}_{n, x}(\sigma^{2}((Decoder(S_{s_{feature_{i}}}, S_{t}) \cdot Mask_{R_{j}})))
\end{equation}
where,
\begin{equation}
Encoder(x) = \{S_{s}, S_{t}\} 
\end{equation}
\begin{equation}
    S_{s} \cdot Mask_{R_{j}} = S_{s_{feature_{j}}}
\end{equation}
In essence, activeness of $S_{s_{feature_{i}}}$ with respect to $R_{j}$ for a given input $x$, is the expectation of variance observed in $R_{j}$ due to addition of noise ($n \in 0.01 \cdot N(0,1)$) to $S_{s_{feature_{i}}}$, taken over all the added noise tensors. We also define the average activeness map ($Map$) for a data distribution, such that $Map_{ij} = \mathbb{E}_{x}(A_{ij})$. \par Figure 3 depicts the average activeness map for our model obtained on test data sampled from the CelebAMask-HQ dataset. 
\begin{figure}[h]
\centering
\includegraphics[scale=0.25]{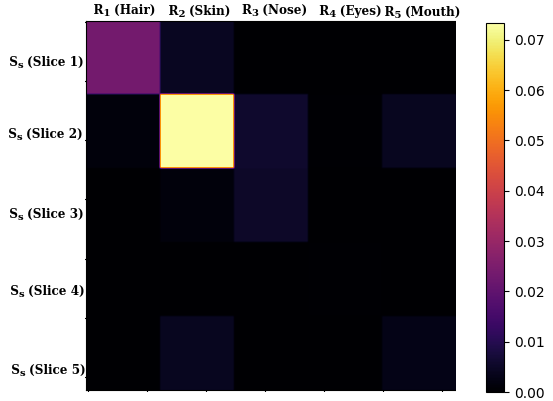}
\caption{Our model's average activeness map obtained on test data sampled from the CelebAMask-HQ dataset. The near diagonal nature of the map suggests a high level of disentanglement among the slices of $S_{s}$ with respect to the intended semantic ROIs.}
\end{figure}
Since the map (matrix) is nearly diagonal, $S_{s_{feature_{i}}}$ controls the pixel intensity levels in $R_{j}$ significantly, only if $j = i$. 
Thus, the claim made 
\begin{table}
\centering
\resizebox{7.5cm}{!}{
\begin{tabular}{|c|c|c|c|}\hline
\textbf{Method} & \textbf{Latent Dimensions} & \textbf{Slicing Scheme} & \textbf{ACS}$\uparrow$ \\
\hline
 &  & [0:4, :] : Hair & \\
 &  & [4:8, :] : Skin & \\
Pixel2Style2Pixel & 18 $\times$ 512 & [8:12, :] : Nose & 0.189\\
\cite{richardson2021encoding} & [18, 512] & [12:16, :] : Eyes & \\
 & & [16:18, :] : Lips + Mouth & \\
\hline
 &  & [0:4, :] : Hair & \\
 &  & [4:8, :] : Skin & \\
StyleGAN2\-ADA & 18 $\times$ 512 & [8:12, :] : Nose & 0.194\\
 \cite{Karras2020ada}& [18, 512] & [12:16, :] : Eyes & \\
 & & [16:18, :] : Lips + Mouth & \\
\hline
 &  & [0:2, :, :] : Hair & \\
 &  & [2:4, :, :] : Skin & \\
Swapping Autoencoder & 8 $\times$ 8  $\times$ 8 & [4:6, :, :] : Nose & 0.259\\
 \cite{park2020swapping}& [8, 8, 8] & [6:7, :, :] : Eyes & \\
 & & [7:8, :, :] : Lips + Mouth & \\
\hline
&  & [0:2, :, :] : Hair & \\
 &  & [2:4, :, :] : Skin & \\
Latents2Segments & 8 $\times$ 8  $\times$ 8 & [4:6, :, :] : Nose & \textbf{0.819}\\
(Ours) & [8, 8, 8] & [6:7, :, :] : Eyes & \\
 & & [7:8, :, :] : Lips + Mouth & \\
\hline
\end{tabular}}
\caption{Comparative analysis of our model's latent space with that of SOTA Generative Models which either encode or project input images onto a structured latent space on the basis of ACS using test images from CelebAMask-HQ \cite{CelebAMask-HQ} dataset. Our model's latent space is the most disentangled and by a large margin, with respect to semantic ROIs.}
\end{table}
in section 2.1 regarding the correspondences that our model develops, stands validated. We define the ratio of sum of diagonal elements of the average activeness map to that of sum of all elements of the average activeness map as the \textbf{\textit{Activeness Compaction Score} (ACS)}. The ACS for our model with regard to the CelebAMask-HQ dataset\cite{CelebAMask-HQ} was found to be \textbf{0.8186} which suggests a nearly diagonal nature of the obtained $Map$. Thus, it is evident that the slices of our model's latent space have a direct correspondence with pixel intensities in the intended semantic regions of interest only. We present a comparative analysis of our model's latent space with that of several SOTA works in Table 1. The slicing schema chosen for these experiments were taken to be close to ours in order to maintain consistency. Our model's latent space has the most disentanglement (with respect to semantic ROIs) infused in it.   
\begin{figure}
    \centering
    \resizebox{8cm}{!}{
      \begin{tabular}{c c c c}
        \includegraphics[scale = 0.25]{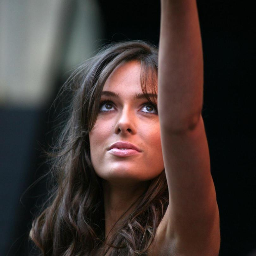} & \includegraphics[scale = 0.5]{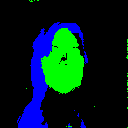} &      
        \includegraphics[scale = 0.125]{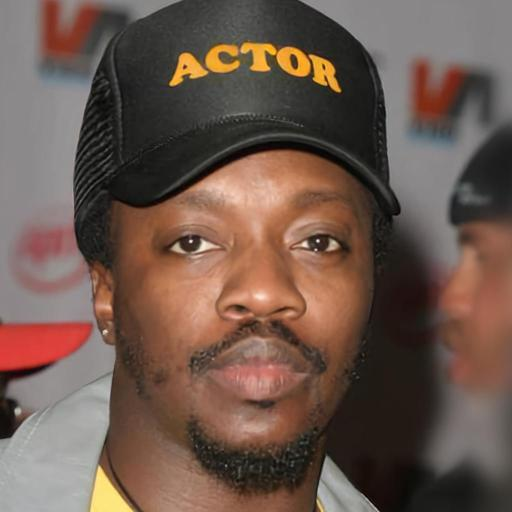} & \includegraphics[scale = 0.5]{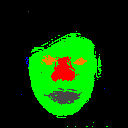}\\
        \multicolumn{2}{c}{(a)} & \multicolumn{2}{c}{(b)}\\  
    \end{tabular}}
    \caption{Challenging instances from test images belonging to (a) HELEN (in the wild) and (b) CelebAMask-HQ dataset. Our model performed well despite the occlusions (regions not belonging to any semantic ROI), being present. The color coding used for the predicted segmentation maps is the same as that used for Figure 1.}
    \label{fig:my_label}
\end{figure}

\begin{table}
\centering
\resizebox{7.5cm}{!}{
\begin{tabular}{|c|c c c c c |c|}
 \hline
 \multirow{2}{*}{\textbf{Method}} & \multicolumn{5}{|c|}{\textbf{F1 Score} $\uparrow$} & \multirow{2}{*}{\textbf{Computation Time(ms)} $\downarrow$}\\
  & \textbf{Hair} & \textbf{Skin} & \textbf{Nose} &\textbf{Eyes} & \textbf{Skin$+$Mouth}&\\ 
 \hline
  Modified BiSeNet & 0.9524 & 0.89 & 0.931 & 0.81 & 0.741 & 142.69\\ 
 \hline
  Ours & 0.7803 & 0.8532 & 0.7605 & 0.578 & 0.6715 & \textbf{124.00}\\
 \hline
\end{tabular}}
\caption{F1 scores and time taken per input image for segmentation with respect to different semantic ROIs obtained on test images from the CelebAMask-HQ dataset. Our model's disentangled latent space yields performance comparable to publicly available pre-trained SOTA (Modified BiSeNet\cite{9577672}) for most ROIs and the rate of inference is faster.}
\end{table}
\subsection{Segmentation Accuracy}
Qualitative results have been presented in Figure 1 (refer to section 1) and Figure 4. We chose to disentangle two large slices (each containing 4 channels) of $S_{s}$, with respect to \textit{Hair} and \textit{Skin}, respectively, while working with the HELEN dataset as the number of training images was lesser than that required to infuse disentanglement within several slices. We compare the accuracy of our model's predicted semantic labels for different ROIs and its rate of inference with  SOTA,  on the basis F1 scores and computation time per input image in Table 2. Our model is faster and comparable to SOTA for the chosen downstream task. From Table 2, we infer that certain entanglements in the latent space of our model are essential for near perfect re-generation of structure information by the decoder, since it inherits from the StyleGAN2\cite{Karras_2020_CVPR} architecture. Therefore, there is a trade-off between the extent to which $S_{s}$ is disentangled, and the segmentation accuracy obtained. Since, this work focuses on disentanglement, we have optimized only the amount of disentanglement achieved, and will take up the characterization of this trade-off as a future work. We also observe that the accuracy of predicted segmentation maps and the number of classes for which segmentation is feasible, has a direct correlation  with the amount of varied and well-annotated training data available.
\section{Conclusion}
In this work, we proposed and evaluated a method to disentangle the latent space of Generative Autoencoder models with respect to semantic ROIs. We illustrated its applicability to the downstream task of semantic segmentation of face images. Our model outperforms SOTA in terms of disentanglement and is faster while being comparably accurate in performing the downstream task. Our model shall find tremendous utility, especially in AR/VR applications that require selective control over semantic ROIs as a prerequisite since it is entirely prior-agnostic.
\vspace{-3mm}
\paragraph{Acknowledgement:}  Support from Institute of Eminence (IoE) project No. SB20210832EEMHRD005001 is gratefully acknowledged.
{\small
\bibliographystyle{ieee_fullname}
\bibliography{egbib}
}
\end{document}